\documentclass[11pt,a4paper]{article}
\usepackage[hyperref]{naaclhlt2019}
\usepackage{times}
\usepackage{latexsym}

\aclfinalcopy 

\usepackage{array}
\usepackage{amsmath}
\usepackage{amssymb}
\usepackage[english]{babel}
\usepackage{color}
\usepackage{multicol}
\usepackage{multirow}
\usepackage{array}
\usepackage{balance}
\usepackage{arydshln}
\usepackage{tikz}
\usepackage{url}
\usepackage{xspace}

\newcommand{\ie}{\emph{i.e.,}\xspace}

\usepackage{helvet}
\usepackage{courier}

\usepackage{bm}

\usepackage{amsfonts}
\usepackage{graphicx} 
\usepackage{graphics}
\usepackage{standalone}

\usepackage{caption}
\usepackage{subfig}
\usepackage{mathrsfs}

\newcommand{\xing}[1]{\textcolor{red}{wx: #1}}

\title{Modeling Recurrence for Transformer}

\author{Jie Hao\\\normalsize Florida State University\\{\em \normalsize haoj8711@gmail.com} \And
Xing Wang\\\normalsize Tencent AI Lab\\{\em \normalsize brightxwang@tencent.com} \And 
Baosong Yang\\\normalsize University of Macau\\{\em \normalsize nlp2ct.baosong@gmail.com}
\AND
Longyue Wang\\\normalsize Tencent AI Lab\\{\em \normalsize vinnylywang@tencent.com} \And 
Jinfeng Zhang\\\normalsize Florida State University\\{\em \normalsize jinfeng@stat.fsu.edu} \And
Zhaopeng Tu\thanks{~~~Zhaopeng Tu is the corresponding author of the paper. This work was conducted when Jie Hao and Baosong Yang were interning at Tencent AI Lab.}\\\normalsize Tencent AI Lab\\{\em \normalsize zptu@tencent.com}
}

\begin{document}
\maketitle
\begin{abstract}
Recently, the Transformer model~\cite{Vaswani:2017:NIPS} that is based solely on attention mechanisms, has advanced the state-of-the-art on various machine translation tasks. However, recent studies reveal that the lack of recurrence
hinders its further improvement of translation capacity~\cite{Chen:2018:ACL,Dehghani:2019:ICLR}. In response to this problem, we propose to directly model recurrence for Transformer with an additional recurrence encoder. In addition to the standard recurrent neural network, we introduce a novel {\em attentive recurrent network} to leverage the strengths of both attention and recurrent networks. Experimental results on the widely-used WMT14 English$\Rightarrow$German and WMT17 Chinese$\Rightarrow$English translation tasks demonstrate the effectiveness of the proposed approach. Our studies also reveal that the proposed model benefits from a {\em short-cut} that bridges the source and target sequences with a single recurrent layer, which outperforms its deep counterpart.

\end{abstract}

\section{Introduction}
Recently, Transformer~\cite{Vaswani:2017:NIPS} -- a new network architecture based solely on attention mechanisms, has advanced the state-of-the-art on various translation tasks across language pairs. 
Compared with the conventional recurrent neural network (RNN) \cite{Schuster:1997:TSP} based model that leverages recurrence as the basic building module~\cite{sutskever2014sequence,Bahdanau:2015:ICLR,Chen:2018:ACL}, Transformer replaces RNN with self-attention network (SAN) to model the dependencies among input elements. One appealing strength of SAN is that it breaks down the sequential assumption to obtain the ability of highly parallel computation: input elements interact with each other simultaneously without regard to their distance.



However, prior studies empirically show that the lack of recurrence modeling hinders Transformer from further improvement of translation quality~\cite{Dehghani:2019:ICLR}. Modeling recurrence is crucial for capturing several essential properties of input sequence, such as structural representations~\cite{Tran:2016:NAACL} and positional encoding~\cite{Shaw:2018:NAACL}, which are exactly the weaknesses of SAN~\cite{Tran:2018:arXiv}.
Recently,~\newcite{Chen:2018:ACL} show that the representations learned by SAN-based and RNN-based encoders are complementary to each other, and merging them can improve translation performance for RNN-based NMT models.



Starting from these findings, we propose to directly model recurrence for Transformer with an additional recurrence encoder. The recurrence encoder recurrently reads word embeddings of input sequence and outputs a sequence of hidden states, which serves as an additional information source to the Transformer decoder.
In addition to the standard RNN, we propose to implement recurrence modeling with a novel {\em attentive recurrent network} (ARN), which combines advantages of both SAN and RNN.
Instead of recurring over the individual symbols of sequences like RNN, the ARN recurrently revises its representations over a set of feature vectors, which are extracted by an attention model from the input sequence. Accordingly, ARN combines the strong global modeling capacity of SAN with the recurrent bias of RNN.
%

We evaluate the proposed approach on widely-used WMT14 English$\Rightarrow$German and WMT17 Chinese$\Rightarrow$English translation tasks.
Experimental results show that the additional recurrence encoder, implemented with either RNN or ARN, consistently improves translation performance, demonstrating the necessity of modeling recurrence for Transformer. Specifically, the ARN implementation outperforms its RNN counterpart, which confirms the strength of ARN. 

Further analyses reveal that our approach benefits from a {\em short-cut} that bridges the source and target sequences with shorter path. Among all the model variants, the implementation with shortest path performs best, in which the recurrence encoder is single layer and its output is only fed to the top decoder layer. It consistently outperforms its multiple deep counterparts, such as multiple-layer recurrence encoder and feeding the output of recurrence encoder to all the decoder layers. In addition, our approach indeed generates more informative encoder representations, especially representative on syntactic structure features, through conducting linguistic analyses on probing tasks~\cite{conneau2018you}.

\section{Background}

\begin{figure}[t]
    \centering
    \includegraphics[width=0.38\textwidth]{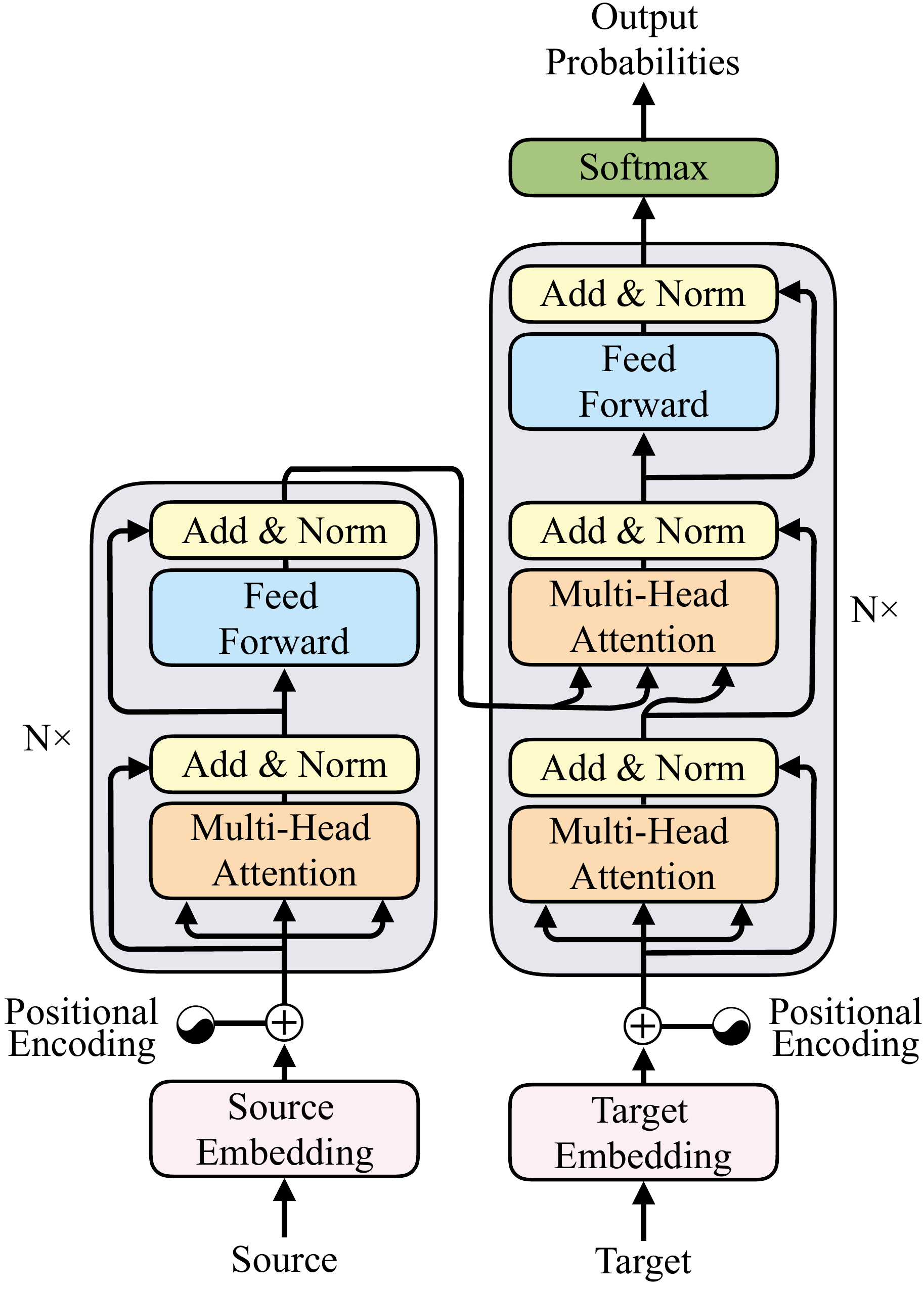}
    \caption{The architecture of Transformer.}
    \label{fig:transformer}
\end{figure}

Figure~\ref{fig:transformer} shows the model architecture of Transformer. 
The encoder is composed of a stack of $N$ identical layers, each of which has two sub-layers. The first sub-layer is a self-attention network, and the second one is a position-wise fully connected feed-forward network. A residual connection~\cite{he2016deep} is employed around each of two sub-layers, followed by layer normalization~\cite{ba2016layer}.
Formally, the output of the first sub-layer ${\bf C}_e^n$ and the second sub-layer ${\bf H}_e^n$ are sequentially calculated as:
\begin{eqnarray}
    {\bf C}_e^n &=& \textsc{Ln}\big(\textsc{Self-Att}({\bf H}_e^{n-1}) + {\bf H}_e^{n-1} \big), \\
    {\bf H}_e^n &=& \textsc{Ln}\big(\textsc{Ffn}({\bf C}_e^n) + {\bf C}_e^{n} \big),
    \label{eqn:enc}
\end{eqnarray}
where $\textsc{Self-Att}(\cdot)$, $\textsc{Ln}(\cdot)$, and $\textsc{Ffn}(\cdot)$ are respectively self-attention mechanism, layer normalization, and feed-forward network with ReLU activation in between. 

In transformer, $\textsc{Self-Att}(\cdot)$ computes attention over the input ${\bf H}_e^{n-1}$ as follows:
\begin{equation}
\textsc{Self-Att}({\bf H}_e^{n-1}) =  softmax(\frac{{\bf Q} {\bf K}^{\top} }{ \sqrt{d_k}}  ) {\bf V}
\end{equation}
where $\{{\bf Q}, {\bf K}, {\bf V}\}$ are query, key and value vectors that are transformed from the  input representations ${\bf H}_e^{n-1}$. $\sqrt{d_k}$ is the scaling factor where the $d_k$ is the dimension size of the query and key vectors. 

The decoder is also composed of a stack of $N$ identical layers. In addition to two sub-layers in each decoder layer, the decoder inserts a third sub-layer ${\bf D}_d^n$ to perform attention over the output of the encoder ${\bf H}^{N}_{e}$:
\begin{eqnarray}
    {\bf C}_d^n &=& \textsc{Ln}\big(\textsc{Self-Att}({\bf H}_d^{n-1}) + {\bf H}_d^{n-1} \big), \label{eqn:dec_c} \\
    {\bf D}_d^n &=& \textsc{Ln}\big(\textsc{Att}({\bf C}_d^{n}, {\bf H}_e^N) + {\bf C}_d^{n} \big), \label{eqn:dec_d} \\
    {\bf H}_d^n &=& \textsc{Ln}\big(\textsc{Ffn}({\bf D}_d^n) + {\bf D}_d^{n} \big) \label{eqn:dec_h},
\end{eqnarray}
where $\textsc{Att}({\bf C}_d^{n}, {\bf H}_e^N)$ denotes attending the top encoder layer ${\bf H}_e^N$ with ${\bf C}_d^n$ as query. 
The top layer of the decoder ${\bf H}_d^N$ is used to generate the final output sequence.

\section{Approach}
In this section, we first describe the architecture of the introduced recurrence encoder and elaborate two types of neural network that are used as recurrence encoder in this work. Then we introduce the integration of recurrence encoder into the Transformer. Specifically, two strategies are presented to fuse the representations produced by the recurrence encoder and the conventional encoder.  Finally we present the {\em short-cut} connection between the recurrence encoder and the decoder that we found very effective to use the learned representation to improve the translation performance under the proposed architecture.

\subsection{Recurrence Modeling}

\begin{figure}[t]
    \centering
     \includegraphics[width=0.45\textwidth]{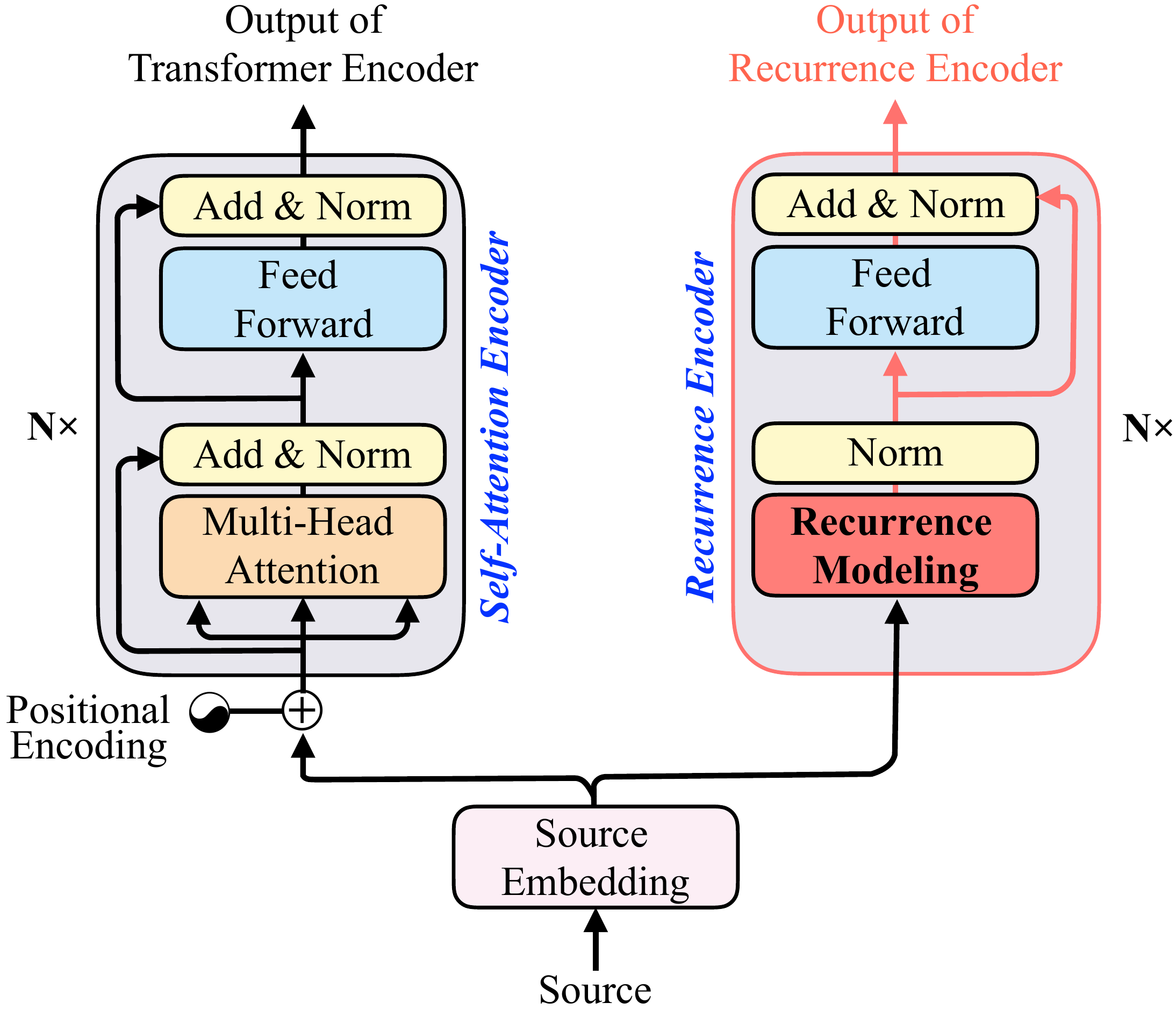}
    \caption{The architecture of Transformer augmented with an additional {\em recurrence encoder}, the output of which is directly fed to the top decoder layer.}
    \label{fig:rec-enc}
\end{figure}

Figure~\ref{fig:rec-enc} shows the architecture of the introduced {\em recurrence encoder} which reads word embeddings of source words and outputs a sequence of hidden states that embeds recurrent information. Similar to the Transformer encoder, it has a stack of $N$ identical layers, each
of which has two sub-layers. The first one is a recurrence modeling network and the second is a fully connected feed-forward network:
\begin{eqnarray}
    {\bf C}_r^{n} &=& \textsc{Ln}(\textsc{Rec}({\bf H}^{n-1}_{r}) + {\bf H}^{n-1}_{r}), \label{eqn:recurrence} \\
    {\bf H}_r^{n} &=& \textsc{Ln}(\textsc{Ffn}({\bf C}_r^{n}) + {\bf C}_r^{n}), \label{eqn:recurrence_ffn}
\end{eqnarray}
where $\textsc{Rec}(\cdot)$ is the function of recurrence modeling. Note that at the bottom layer of the recurrence encoder ($N$=1), we do not employ a residual connection on the recurrence sub-layer (i.e. Equation~\ref{eqn:recurrence}), which releases the constraint that ${\bf C}^{1}_r$ should share the same length with input embeddings sequence ${\bf E}_{in}$\footnote{The input of the lowest layer in the recurrence encoder is the word embeddings of input sequence ${\bf E}_{in}$.}. This offers a more flexible choice of the recurrence functions.

\begin{figure}[t]
    \centering
    \subfloat[Recurrent Neural Network]{ \includegraphics[width=0.35\textwidth]{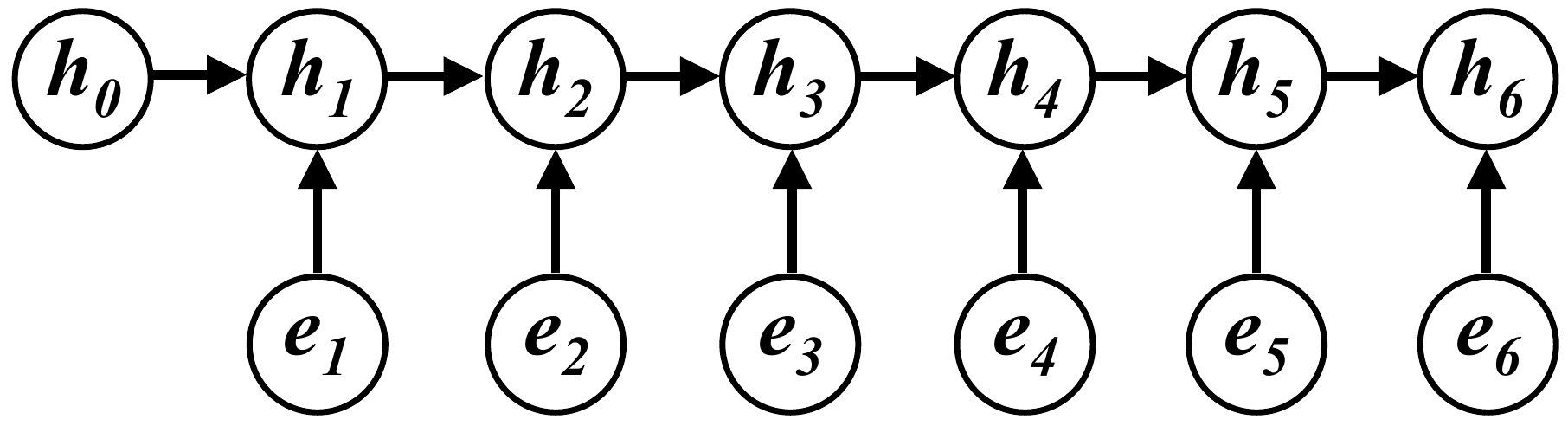} } \\
    \subfloat[Attentive Recurrent Network]{ \includegraphics[width=0.35\textwidth]{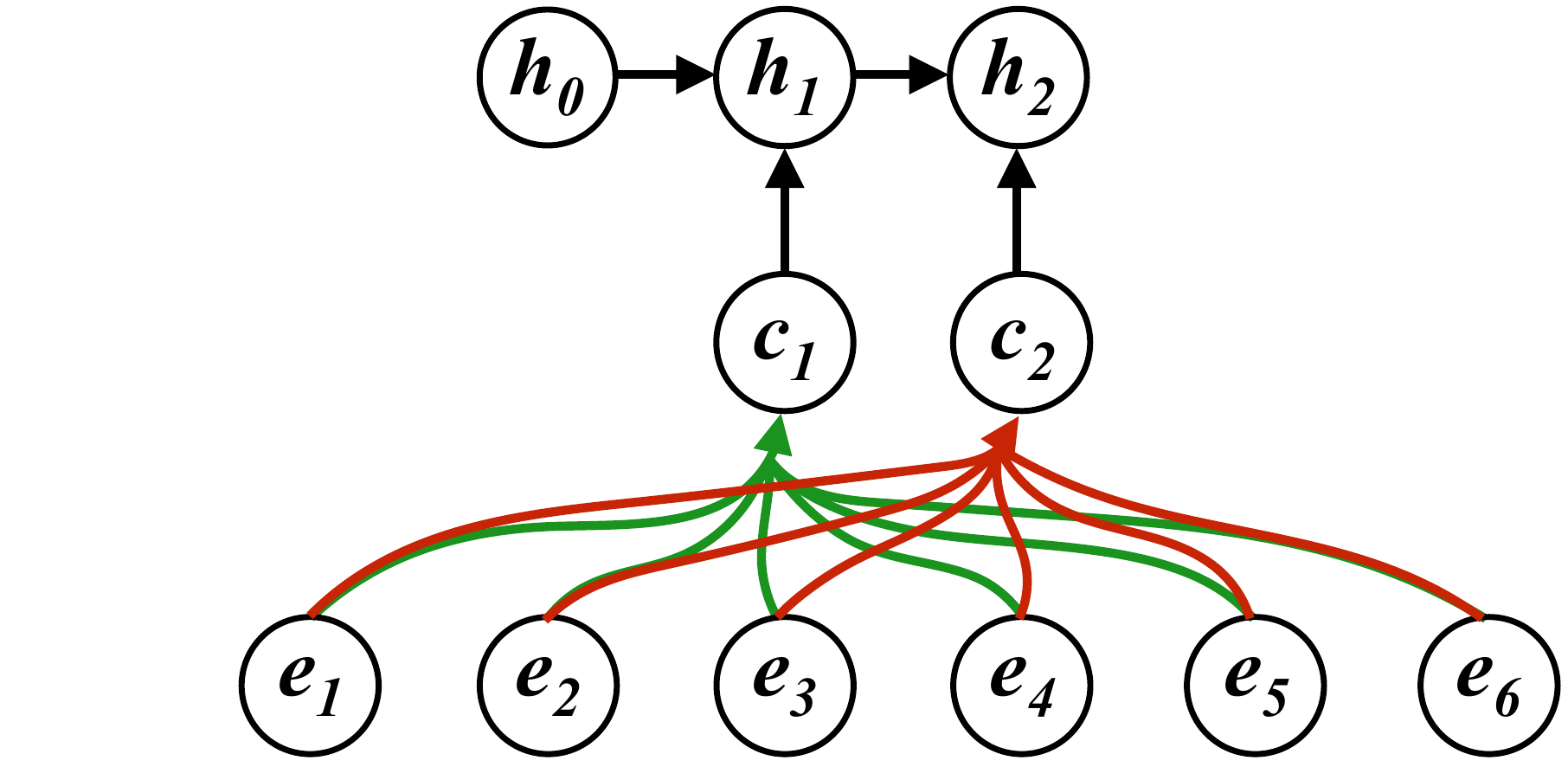} }
    \caption{Two implementations of recurrence modeling: (a) standard RNN, and (b) the proposed ARN.}
    \label{fig:rec-model}
\end{figure}

There are many possible ways to implement the general idea of recurrence modeling $\textsc{Rec}(\cdot)$. The aim of this paper is not to explore this whole space but simply to show that some fairly straightforward implementations work well. 
In this work, we investigate two representative implementations, namely RNN and its variation {\em attentive current network} that combines advantages of both RNN and attention models, as shown in Figure~\ref{fig:rec-model}.


\paragraph{Recurrent Neural Network (RNN)}
An intuitive choice of recurrence modeling is RNN, which is a standard network to model sequence orders. In this work, we use a bidirectional RNN (BiRNN), which is widely applied in RNN-based NMT models~\cite{Bahdanau:2015:ICLR,Chen:2018:ACL}.
Each hidden state in the output representations ${\bf H}^{n}_{\textsc{Rnn}} = \{{\bf h}^{n}_1, \dots, {\bf h}^{n}_J\}$ is calculated as
\begin{eqnarray}
    {\bf h}^{n}_j &=& \big[\overrightarrow{\bf h}_j;  \overleftarrow{\bf h}_j \big] \label{eqn:rnn_concat},\\
    \overrightarrow{\bf h}_j &=& \overrightarrow{f} (\overrightarrow{\bf h}_{j-1}, {\bf h}^{n-1}_j), \\
    \overleftarrow{\bf h}_j &=& \overleftarrow{f} (\overleftarrow{\bf h}_{j+1}, {\bf h}^{n-1}_j), \label{eqn:rnn_back}
\end{eqnarray}
where $\overrightarrow{f}(\cdot)$ and $\overleftarrow{f}(\cdot)$ are the activation functions of forward and backward RNN respectively, which can be implemented as LSTM~\cite{hochreiter1997long} or GRU~\cite{cho2014learning}. ${\bf h}_0^n$ is the initial state of RNN, which is the mean of ${\bf H}_{\textsc{Rnn}}^{n-1}$. ${\bf H}_{\textsc{Rnn}}^{0}$ represents the word embeddings of the input sequence.

\begin{figure*}[t]
    \centering
    \subfloat[Gated Sum]{ \includegraphics[width=0.35\textwidth]{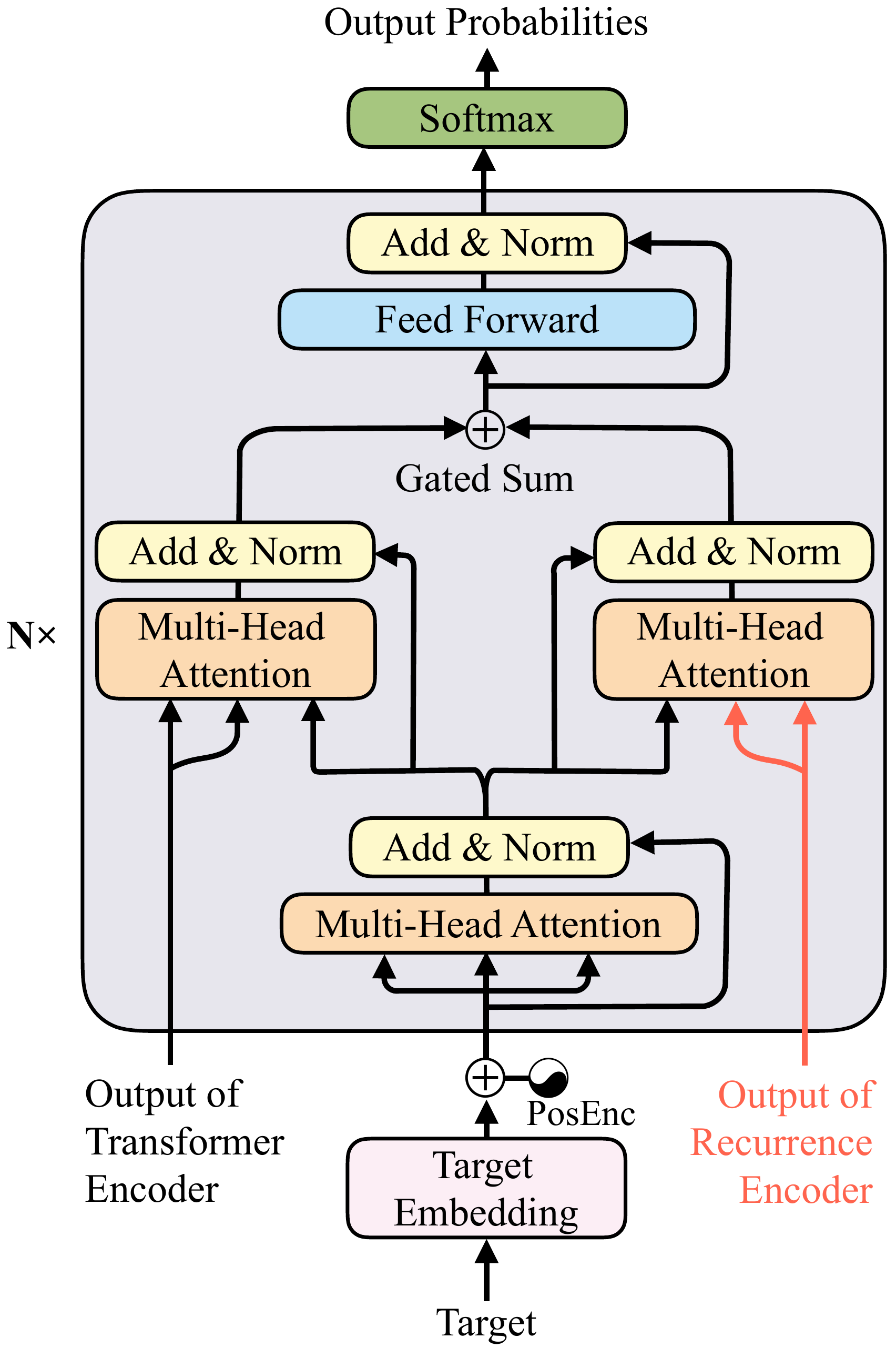} } \hspace{0.15\textwidth}
    \subfloat[Stack]{ \includegraphics[width=0.35\textwidth]{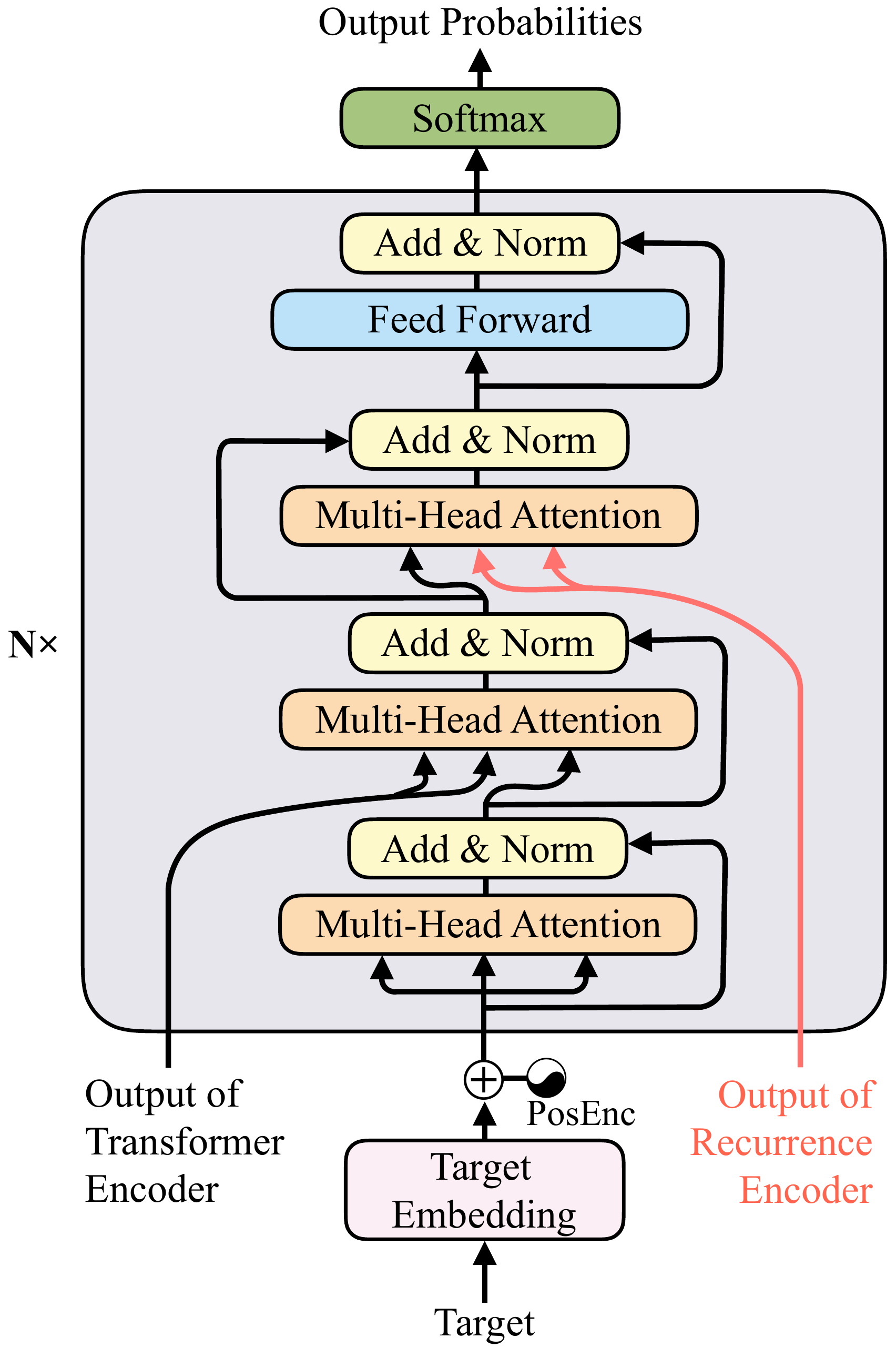} }
    \caption{Different strategies to integrate the output of the additional {\em recurrence encoder} into the decoder.}
    \label{fig:recurrence}
\end{figure*}

\paragraph{Attentive Recurrent Network (ARN)}
We can also extend RNN by recurring over a set of feature vectors extracted with an attention model, which allows the model to learn a compact, abstractive feature vectors over the input sequence.
Specifically, the ARN performs $T$ recurrent steps on the attentive output of the 
input representation ${\bf H}^{n-1}_{r}$:
\begin{eqnarray}
    {\bf h}^{n}_t &=& f({\bf h}^{n}_{t-1}, {\bf c}^{n}_t), \\
    {\bf c}^{n}_t &=& \textsc{Att}({\bf h}^{n}_{t-1}, {\bf H}^{n-1}_{r}).
\end{eqnarray}
The output representations ${\bf H}^n_{\textsc{Arn}} = \{{\bf h}^n_1, \dots, {\bf h}^n_T\}$ are fed to the subsequent modules. Analogous to Equations \ref{eqn:rnn_concat}-\ref{eqn:rnn_back}, ARN can be extended to the bidirectional variant, i.e. BiARN, except that the input is the attentive context vector ${\bf c}^{n}_t$ rather than the individual representation vector of the input sequence.

Note that, the number of recurrence step $T$ is allowed to be unequal to the length of input sequence $J$. 
In contrast to RNN which recurs over the individual symbols of the input sequences, ARN recurrently revises its representations of all symbols in the sequence with an attention model. 


\subsection{Integrating into Transformer}

Since the output of recurrence encoder unnecessarily shares the same length with that of Transformer encoder (e.g. when ARN is used as recurrence function), combination strategy on the encoder side, such as concatenating the outputs of both encoders~\cite{Chen:2018:ACL}, is not an universal solution in this scenario. Accordingly, we feed the information of the additional recurrence encoder into the decoder of Transformer. Specifically, 
we serve an additional attention layer ${\bf R}_d^n$ as the fourth sub-layer in each decoder block to perform attention over the output of the recurrence encoder ${\bf H}^N_r$. As shown in Figure~\ref{fig:recurrence}, we present two strategies to integrate ${\bf R}_d^n$, namely {\em gated sum} and {\em stack}, which differ at how ${\bf R}_d^n$ interacts with the output of attention over the Transformer encoder, i.e., ${\bf D}_d^n$ in Equation~\ref{eqn:dec_d}.


\paragraph{Gated Sum}
The first strategy combines the outputs of the two attention sub-layers in a gating fusion (Figure~\ref{fig:recurrence}(a)), in which the outputs of both encoders are attended simultaneously:
\begin{eqnarray}
    {\bf R}_d^n &=& \textsc{Ln}\big(\textsc{Att}({\bf C}_d^{n}, {\bf H}^N_r) + {\bf C}_d^{n} \big), \label{eqn:dec_gate_r}\\
    \widehat{\bf D}_d^n &=& \lambda_n {\bf D}_d^n + (1-\lambda_n) {\bf R}_d^n, \label{eqn:dec_gate_combine} \\
    {\bf H}_d^n &=& \textsc{Ln}\big(\textsc{Ffn}(\widehat{\bf D}_d^n) + \widehat{\bf D}_d^{n} \big), \label{eqn:dec_gate_h}
\end{eqnarray}
where $\lambda_n$ is an interpolation weight calculated by a logistic sigmoid function:
\begin{equation}
    \lambda_n = sigmoid({\bf D}_d^n, {\bf R}_d^n)
\end{equation}
As seen, the output of self-attention layer ${\bf C}_d^n$ serves as a query to attend the outputs of both encoders (Equations~\ref{eqn:dec_d} and~\ref{eqn:dec_gate_r}), and the outputs of both attention models $\{{\bf D}_d^n, {\bf R}_d^n\}$ are combined via a gated sum (Equation~\ref{eqn:dec_gate_combine}), which is subsequently fed to the feed-forward layer (Equation~\ref{eqn:dec_gate_h}).

\paragraph{Stack}
We can also arrange the sub-layers in a {\em stack} (Figure~\ref{fig:recurrence}(b)), in which the outputs of both encoders are attended sequentially:
\begin{eqnarray}
    {\bf R}_d^n &=& \textsc{Ln}\big(\textsc{Att}({\bf D}_d^{n}, {\bf H}^N_r) + {\bf D}_d^{n} \big), \label{eqn:dec_cascade_r}\\
    {\bf H}_d^n &=& \textsc{Ln}\big(\textsc{Ffn}({\bf R}_d^n) + {\bf R}_d^{n} \big), \label{eqn:dec_cascade_h}
\end{eqnarray}
The decoder first attends the output of Transformer encoder, and the attention output ${\bf D}_d^n$ serves as the query to attend the output of recurrence encoder (Equation~\ref{eqn:dec_cascade_r}).

\subsection{Short-Cut Effect}

The introduced recurrence encoder provides an additional computation path ranging from the input sequence to the output sequence.~\citeauthor{Chung:2017:ICLR}~\shortcite{Chung:2017:ICLR} and~\citeauthor{Shen:2019:ICLR}~\shortcite{Shen:2019:ICLR} have  shown that a shortcut for gradient back-propagation benefits language modeling. Inspired from them, we use a shorter path to transform the learned recurrence. We call this the ``{\em short-cut effect}''.

Among all the model variants, we implement shortest path as: the recurrence encoder is single layer and its output is only fed to the top decoder layer while the first $N-1$ decoder layers perform the same as the standard Transformer (e.g. Equations~\ref{eqn:dec_c}-\ref{eqn:dec_h}). Accordingly, the computation path is ${\bf E}_{in} \rightarrow {\bf H}_r \rightarrow {\bf R}_d^N \rightarrow {\bf H}_d^N$, then the decoder uses ${\bf H}_d^N$ to make a target word prediction. It is much simpler than that of the conventional Transformer, which transfers information learned from input sequences across multiple stacking encoder and decoder layers. We expect it outperforms its multiple deep counterparts, such as multiple-layer recurrence encoder and feeding the output of recurrence encoder to all the decoder layers.

\section{Related Work}

\paragraph{Improving Transformer Encoder}

From the perspective of representation learning, there has been an increasing amount of work on improving the representation power of SAN encoder. \newcite{Bawden:2018:NAACL} and \newcite{Voita:2018:ACL} exploit external context for SAN encoder, while~\newcite{Yang:2019:AAAI} leverage the intermediate representations to contextualize the transformations in SAN.
A number of recent efforts have explored ways to improve multi-head SAN by encouraging individual attention heads to extract distinct information~\cite{Strubell:2018:EMNLP,Li:2018:EMNLP}.
Concerning multi-layer SAN encoder,~\newcite{Dou:2018:EMNLP,Dou:2019:AAAI} and~\newcite{wang:2018:COLING} propose to aggregate the multi-layer representations, and~\newcite{Dehghani:2019:ICLR} recurrently refine these representations. 
Our approach is complementary to theirs, since they focus on improving the representation power of SAN encoder, while we aim to complement SAN encoder with an additional recurrence encoder. 



Along the direction of modeling recurrence for SAN,
~\newcite{Vaswani:2017:NIPS} and~\newcite{Shaw:2018:NAACL} inject absolute position encoding and relative positional encoding to consider the position information respectively.~\newcite{Shen:2018:AAAI} introduce a directional self-attention network (DiSAN), which allows each token to attend to previous (or following) tokens only. Both studies verify the necessity of modeling recurrence for SAN.
We re-implemented these approaches on top of Transformer, and experimental results show that our approach outperforms them by explicitly augmenting Transformer with an additional recurrence encoder. It should be emphasized that our approach is complementary to theirs, and combining them together is expected to further improve performance, which we leave for future work.

Closely related to our work,~\newcite{Chen:2018:ACL} propose to combine SAN encoder with an additional RNN encoder.
The main differences between our work and theirs are: 1) we enhance the state-of-the-art Transformer with recurrence information, while~\newcite{Chen:2018:ACL} augment RNN-based models with SAN encoder. To this end, we propose a novel attentive recurrent network to implement the additional recurrence encoder in Transformer. 
We re-implemented the approach proposed by~\newcite{Chen:2018:ACL} on top of Transformer. Experimental results indicate the superiority of our approach, which confirms our claim.
In addition, we elaborately design the integration strategy to effectively feed the recurrence information to the decoder, and empirically show that the proposed model benefits from the {\em short-cut} effect.


\paragraph{Comparison to Reviewer Network}
Attentive recurrent network are inspired by the reviewer network, which is proposed by~\newcite{Yang:2016:NIPS} for the image caption generation task. There are two key differences which reflect how we have generalized from the original model. First, we perform attention steps over the source embeddings instead of the encoder representations. The main reason is that the Transformer encoder is implemented as multiple layers, and higher layers generally encode global information, as indicated by~\citeauthor{Peters:2018:NAACL}~\shortcite{Peters:2018:NAACL}.
Second, we feed the feature vectors together with the original encoder representations to the decoder. In image caption generation, the source side (i.e. image) contains much more information than the target side (i.e. caption)~\cite{Tu:2017:TACL}. Therefore, they aim at learning a compact and abstractive representation from the source information, which serves as the only input to the decoder. 
In this work, we focus on leveraging the attention model to better learn the recurrence, which we expect to complement the Transformer model.
In our preliminary experiments, attending over the encoder representations does not improve performance, while feeding the feature vectors only to the decoder seriously harms performance.

\section{Experiment}

\subsection{Setup}
We conducted experiments on the widely-used WMT14 English-to-German (4.6M sentence pairs, En$\Rightarrow$De) and WMT17 Chinese-to-English (20.6M sentence pairs, Zh$\Rightarrow$En) translation tasks.
All the data had been tokenized and segmented into subword symbols using byte-pair encoding \cite{sennrich2016neural} with 32K merge operations\footnote{https://github.com/rsennrich/subword-nmt}.
We used case-sensitive NIST BLEU score \cite{papineni2002bleu} as the evaluation metric, and {\em bootstrap resampling} \cite{koehn2003statistical} for statistical significance test.

We implemented the proposed approaches on top of the Transformer model~\cite{Vaswani:2017:NIPS}. Both in our model and related model of Subsection 5.3, the RNN is implemented with GRU~\cite{cho2014learning} for fair comparison.
We followed the configurations in \newcite{Vaswani:2017:NIPS}, and reproduced their reported results on the En$\Rightarrow$De task. 

We initialized parameters of the proposed models by the pre-trained baseline model.  We have tested both {\em Base} and {\em Big} models, which differ at hidden size (512 vs. 1024), filter size (2048 vs. 4096), and number of attention heads (8 vs. 16). 
In consideration of computation cost, we studied model variations with {\em Base} model on En$\Rightarrow$De task, and evaluated overall performances with both {\em Base} and {\em Big} models on both En$\Rightarrow$De and Zh$\Rightarrow$En translation tasks. 

\subsection{Impact of Components}
In this subsection, we conducted ablation studies to evaluate the different implementations of the proposed model, e.g., recurrence encoder and integration strategy, under the proposed architecture.

\paragraph{Effect of Recurrence Modeling}
We first investigated the effect of recurrence encoder implementations, as listed in Table~\ref{table:recurrence}. We observed that introducing an additional recurrence encoder improves translation performance in all cases. Among all model variations, \textsc{BiARN} outperforms its \textsc{BiRNN} counterpart.

Concerning \textsc{BiARN} models, reducing the layers consistently improves performance. Specifically, the 1-Layer \textsc{BiARN} achieves the best performances in both translation quality and training speed.
This confirms the claim that the proposed approach benefits from a {\em short-cut} on gradient back-propagation.
Accordingly, we adopted 1-Layer \textsc{BiARN} as the default setting in the following experiments.

\begin{table}[t]
\centering
    \begin{tabular}{l| l || c | c}
    {\bf Model} &   \bf Rec. Encoder & \bf Speed  &{\bf BLEU}\\
    \hline \hline
    \textsc{Base}   &   n/a & 1.28 & 27.31\\
    \hline
    \multirow{4}{*}{\textsc{Ours}}   
        & \small  6-Layer    \textsc{BiRNN} & 1.10 & 27.54 \\
        \cdashline{2-4}
        & \small 6-Layer \textsc{BiARN} & 1.09 & 27.72     \\
        & \small 3-Layer \textsc{BiARN} & 1.15 & 28.10     \\
        & \small 1-Layer \textsc{BiARN} & 1.24 &  \bf 28.21 \\ 
    \end{tabular}
  \caption{Evaluation of recurrence encoder implementations. The output of recurrence encoder is fed to the top decoder layer in a stack fusion. 
  ``Speed" denotes the training speed (steps/second).} 
  \label{table:recurrence}
\end{table}

\begin{table}[t]
\centering
\begin{tabular}{l| l | c || c}
    {\bf Model} &   \bf Integration &   \bf to Dec. & {\bf BLEU}\\
    \hline \hline
    \textsc{Base}   &   n/a &   n/a & 27.31\\
    \hline
    \multirow{4}{*}{\textsc{Ours}}  
        &   Gated Sum   &   Top &  28.12 \\
        &   Gated Sum   &   All &  28.02  \\
    \cdashline{2-4}
        &   Stack     &   Top &   \bf 28.21\\
        &   Stack     &   All &   27.93\\
    \end{tabular}
  \caption{Evaluation of decoder integration strategies.} 
  \label{table:integration}
\end{table}

\begin{table*}[t]
\begin{center}
\begin{tabular}{l|l||rl|rl}
    \multirow{2}{*}{\bf System}  &   \multirow{2}{*}{\bf Architecture}  & \multicolumn{2}{c}{\bf Zh$\Rightarrow$En}  &  \multicolumn{2}{|c}{\bf En$\Rightarrow$De}\\
    \cline{3-6}
        &   &   \# Para. &   BLEU    &   \# Para.   &   BLEU\\
    \hline \hline
    \multicolumn{6}{c}{{\em Existing NMT systems}} \\
    \hline
    \multirow{2}{*}{\small  \cite{Vaswani:2017:NIPS}} &   \textsc{Transformer-Base}    &    n/a & n/a &  65M &   27.3\\ 
    &  \textsc{Transformer-Big}               &  n/a  &  n/a  &  213M &  28.4\\ 
    \hdashline
    \small \cite{hassan2018achieving}  &   \textsc{Transformer-Big}  &  n/a  &  24.2  &  n/a  & n/a\\
    \hdashline
    \small  \cite{Chen:2018:ACL}        &  \textsc{Rnmt} + \textsc{San} Encoder 
    &  n/a  &  n/a  &  n/a  & 28.84\\    
    \hline\hline
    \multicolumn{6}{c}{{\em Our NMT systems}}   \\ \hline
    \multirow{5}{*}{\em this work}  &   \textsc{Transformer-Base}  &    107.9M  & 24.13  &  88.0M  &   27.31\\
    &   ~~~ +  1-Layer \textsc{BiARN}                & +9.4M  & 24.70$^\Uparrow$  & +9.4M  & 28.21$^\Uparrow$  \\ 
    \cline{2-6}
    &   \textsc{Transformer-Big}                 & 303.9M  &   24.56 &  264.1M &  28.58   \\ 
    &   ~~~ +  1-Layer \textsc{BiARN}               & +69.4M  &  25.10$^\Uparrow$  & +69.4M & 28.98$^\uparrow$ \\ 
  \end{tabular}
  \caption{Comparing with the existing NMT systems on WMT17 Zh$\Rightarrow$En and WMT14 En$\Rightarrow$De test sets.  ``$\uparrow/\Uparrow$'': significant over the conventional self-attention counterpart ($p < 0.05/0.01$), tested by bootstrap resampling.} 
  \label{tab:exist}
  \end{center}
\end{table*}

\paragraph{Effect of Integration Strategies}
We then tested the effect of different integration strategies, as showed in Table~\ref{table:integration}. We have two observations. First, feeding only to the top decoder layer consistently outperforms feeding to all decoder layers with different integration strategies. This empirically reconfirms the short-cut effect. Second, the stack strategy marginally outperforms its gated sum counterpart.
Therefore, in the following experiments, we adopted the ``Stack + Top'' model in Table~\ref{table:integration} as defaulting setting.

\subsection{Results}

\paragraph{Performances across Languages}
Finally, we evaluated the proposed approach on the widely used WMT17 Zh$\Rightarrow$En and WMT14 En$\Rightarrow$De data, as listed in Table~\ref{tab:exist}. 

To make the evaluation convincing, we reviewed the prior reported systems, and built strong baselines which outperform the reported results on the same data.
As seen in Table~\ref{tab:exist}, modeling recurrence consistently improves translation performance across model variations (\textsc{Base} and \textsc{Big} models) and language pairs (Zh$\Rightarrow$En and En$\Rightarrow$De), demonstrating the effectiveness and universality of our approach. 

\paragraph{Comparison with Previous Work}
\begin{table}[t]
\begin{center}
\begin{tabular}{l| c }
    {\bf Model} & {\bf BLEU}  \\
    \hline \hline
    \textsc{Transformer-Base} & 27.31 \\
    \hline
    ~ + \textsc{RelPos}  & 27.64  \\
    ~ + \textsc{DiSAN}  & 27.58  \\
    \hline
    ~ + \textsc{RNN} Encoder &   27.47 \\
    \hline
    ~ + \textsc{BiARN} Encoder (\textsc{Ours})        &   \bf 28.21     \\
    \end{tabular}
  \caption{Comparison with re-implemented related work: ``\textsc{RelPos}'': relative position encoding~\cite{Shaw:2018:NAACL}, ``\textsc{DiSAN}'': directional SAN~\cite{Shen:2018:AAAI}, ``\textsc{RNN} Encoder'': combining SAN and RNN encoders with multi-column strategy~\cite{Chen:2018:ACL}. }
  \label{table:comparison}
    \end{center}
\end{table}

In order to directly compare our approach with the previous work on modeling recurrence, we re-implemented their approaches on top of the \textsc{Transformer-base} in WMT14 En$\Rightarrow$De translation task.  For relative position encoding, we used unique edge representations per layer and head with clipping distance $k=16$. For the DiSAN strategy, we applied a mask to the \textsc{Transformer} encoder, which constrains the SAN to focus on forward or backward elements. For the multi-column encoder, we re-implemented the additional encoder with six RNN layers.

Table~\ref{table:comparison} lists the results. As seen, all the recurrence enhanced approaches achieve improvements over the baseline model \textsc{Transformer-base}, which demonstrates the necessity of modeling recurrence for \textsc{Transformer}. Among these approaches, our approach (\ie 1-Layer \textsc{BiARN} Encoder) achieves the best performance. 


\subsection{Analysis}

\begin{table*}[t]
  \centering
  \renewcommand{\arraystretch}{1.2} 

  \scalebox{0.92}{
  \begin{tabular}{c||c c | c c c | c c c c c}
    \multirow{2}{*}{\bf Model}   &  \multicolumn{2}{c|}{\bf Surface} & \multicolumn{3}{c|}{\bf Syntactic} &     \multicolumn{5}{c}{\bf Semantic}\\
    \cline{2-11}
     &\bf{SeLen} & \bf {WC} & \bf {TrDep} & \bf {ToCo} & \bf {BShif} & \bf {Tense} & \bf {SubN} & \bf {ObjN} & \bf {SoMo} & \bf {CoIn}\\
    \hline
    \hline
    \multirow{1}{*}{\textsc{Base}} 
     & \bf 92.20 & 63.00 & 44.74 & 79.02 & 71.24 & \bf 89.24 & 84.69 & 84.53 & 52.13 & 62.47\\
    \hline
    \multirow{1}{*}{6-Layer \textsc{BiRNN}} 
      & 89.90 & \bf 77.46 & 44.47 & 79.55 & 71.53 & 89.17 & \bf 85.99 & 84.96 & 51.75 & 61.92 \\
     \multirow{1}{*}{6-Layer \textsc{BiARN}} 
       & 89.78 & 72.02 & 44.45 & 79.21 & 71.31 & 88.38 & 85.64 & \bf 85.00 & 53.27 & 62.38 \\
     \multirow{1}{*}{3-Layer \textsc{BiARN}} 
       & 89.80 & 72.61 & 44.28 & 79.43 & 71.84 & 88.93 & 85.79 & 84.99 & 53.30 & 62.42 \\
     \multirow{1}{*}{1-Layer \textsc{BiARN}} 
       & 90.91 & 73.68 & \bf 45.15 & \bf 79.62 & \bf 72.21 & 89.00 & 85.54 & 84.54 & \bf 53.44 & \bf 62.71 \\
  \end{tabular}
  }
  
  \caption{Classification accuracies on 10 probing tasks of evaluating linguistics embedded in the encoder outputs.} 
  \label{tab:probing}
\end{table*}

\paragraph{Effect of Recurrent Steps}

\begin{figure}[h]
    \centering
    \includegraphics[width=0.38\textwidth]{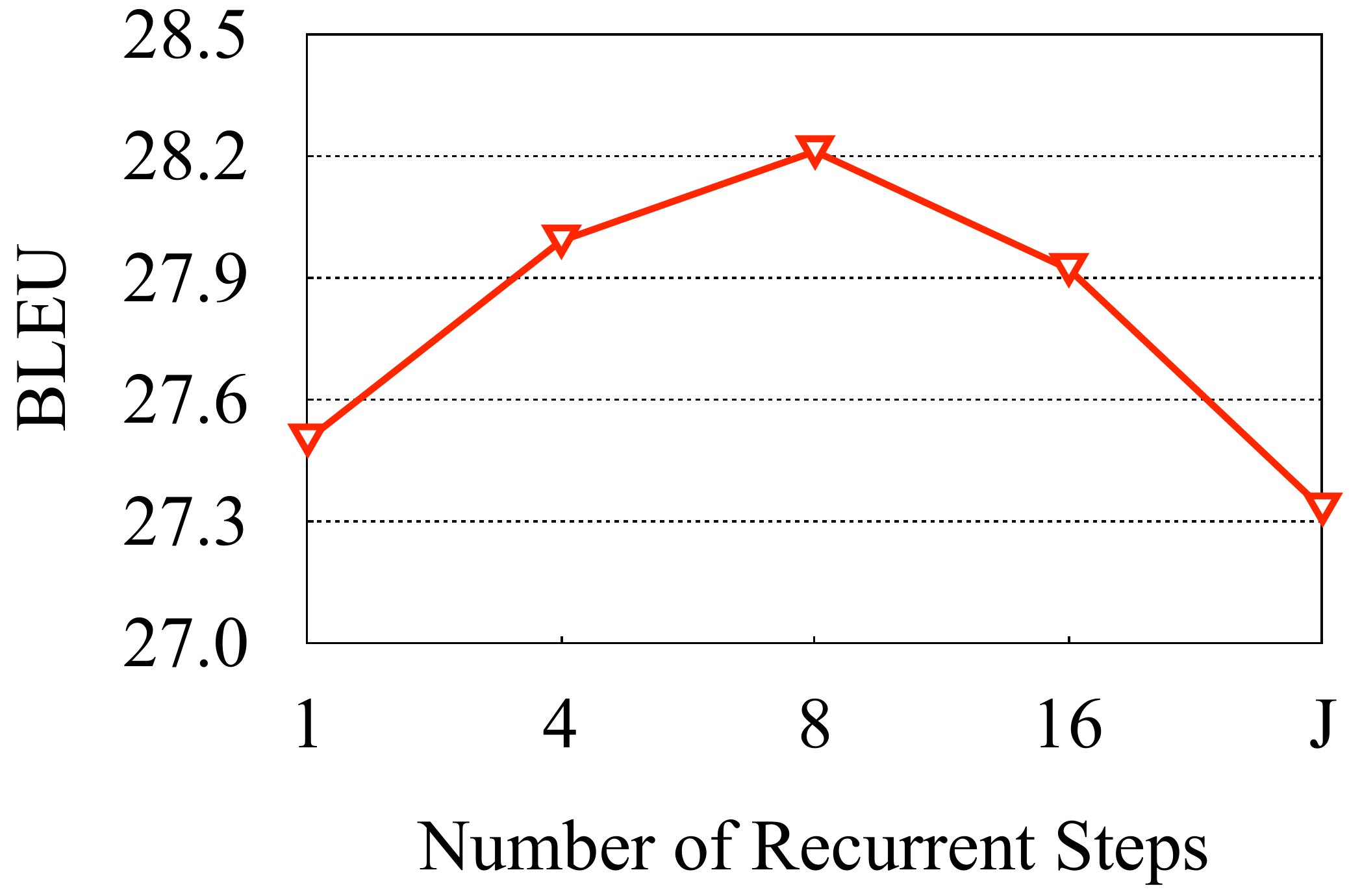}
    \caption{Effect of recurrent steps. The recurrence encoder is implemented as a single-layer {\textsc{BiARN}}. $J$ denotes the length of the input sequence.}
    \label{fig:step}
\end{figure}

To verify the recurrence effect on the proposed model, we conducted experiments with different recurrent steps on single-layer \textsc{BiARN} model.
As shown in Figure~\ref{fig:step}, the BLEU score typically goes up with the increase of the recurrent steps, while the trend does not hold when $T > 8$.
This finding is consistent with~\newcite{Yang:2016:NIPS}, which indicates that conducting too many recurrent steps fails to generate a compact representation. This is exactly one of the ARN's strengths.

\paragraph{Linguistic Analyses}
In this section, we conducted 10 probing tasks\footnote{https://github.com/facebookresearch/SentEval/tree/master\\/data/probing} to study what linguistic properties are captured by the encoders~\cite{conneau2018you}.
A probing task is a classification problem that focuses on simple linguistic properties of sentences. `SeLen' predicts the length of sentences in terms of number of words. `WC' tests whether it is possible to recover information about the original words given its sentence embedding. `TrDep' checks whether an encoder infers the hierarchical structure of sentences. In `ToCo' task, sentences should be classified in terms of the sequence of top constituents immediately below the sentence node. `BShif' tests whether two consecutive tokens within the sentence have been inverted. `Tense' asks for the tense of the main-clause verb. `SubN' focuses on the number of the main clause's subject. `ObjN' tests for the number of the direct object of the main clause. In `SoMo', some sentences are modified by replacing a random noun or verb with another one and the classifier should tell whether a sentence has been modified. `CoIn' contains sentences made of two coordinate clauses. Half of sentences are inverted the order of the clauses and the task is to tell whether a sentence is intact or modified.

We used the pre-trained encoders of model variations in Table~\ref{table:recurrence} to generate the sentence representations of input, which are used to carry out probing tasks. For the \textsc{Transformer-Base} model, the mean of the encoder top layer representations is used as the sentence representation. For the proposed models, which have two encoders, two sentence representations are generated from the same way in base model. To make full use of the learned representations, we combined these two sentence representations via a gate as the final sentence representation to conduct the experiments.

Table~\ref{tab:probing} lists the results. Clearly, the proposed models significantly improve the classification accuracies, although there is still considerable difference among different variants. More specifically,
\begin{itemize}
    \item Concerning surface properties, among the ARN variants, multi-layer ARN inversely decreases the accuracies, while 1-layer ARN consistently improves the accuracies. Considering the related results presented in Table~\ref{table:recurrence} (Row 3-5), we believe that ARN benefits from the shallow structure. 
    \item ARN tends to capture deeper linguistic properties, both syntactic and semantic. Especially, 
    among these probing tasks, `TrDep' and `Toco' tasks are related to syntactic structure modeling.
    As expected, \textsc{Transformer} augmented with an additional encoders outperforms the baseline model, which demonstrates that the proposed models successfully model the syntactic structure.
\end{itemize}

\section{Conclusion}
In this work, we propose to directly model recurrence for Transformer with an additional recurrence encoder. We implement the recurrence encoder with a  novel attentive recurrent network as well as RNN. The recurrence encoder is used to generate recurrence representations for the input sequence. To effectively feed the recurrence representations to the decoder to guide the output sequence generation,  we study two strategies to integrate the recurrence encoder into the Transformer. To evaluate the effectiveness of the proposed model, we conduct experiments on large-scale WMT14 EN$\Rightarrow$DE and WMT17 ZH$\Rightarrow$EN datasets. Experimental results on two language pairs show that the proposed model achieves significant improvements over the baseline \textsc{Transformer}. Linguistic analyses on probing tasks further show that our model indeed generates more informative representations, especially representative on syntactic structure features. 

Future work includes validating the proposed model in other tasks, such as reading comprehension, language inference, and sentence classification. Another promising direction is to directly augment Transformer encoder on recurrence modeling without the additional encoder. 

\section*{Acknowledgments}
J.Z. was supported by the National Institute of General Medical Sciences of the National Institute of Health under award number R01GM126558. 
We thank the anonymous reviewers for their insightful comments.



\balance

\bibliography{main}

\begin{thebibliography}{32}
\expandafter\ifx\csname natexlab\endcsname\relax\def\natexlab#1{#1}\fi

\bibitem[{Ba et~al.(2016)Ba, Kiros, and Hinton}]{ba2016layer}
Jimmy~Lei Ba, Jamie~Ryan Kiros, and Geoffrey~E Hinton. 2016.
\newblock Layer normalization.
\newblock \emph{arXiv preprint arXiv:1607.06450}.

\bibitem[{Bahdanau et~al.(2015)Bahdanau, Cho, and Bengio}]{Bahdanau:2015:ICLR}
Dzmitry Bahdanau, Kyunghyun Cho, and Yoshua Bengio. 2015.
\newblock Neural machine translation by jointly learning to align and
  translate.
\newblock In \emph{ICLR}.

\bibitem[{Bawden et~al.(2018)Bawden, Sennrich, Birch, and
  Haddow}]{Bawden:2018:NAACL}
Rachel Bawden, Rico Sennrich, Alexandra Birch, and Barry Haddow. 2018.
\newblock {Evaluating Discourse Phenomena in Neural Machine Translation}.
\newblock In \emph{NAACL}.

\bibitem[{Chen et~al.(2018)Chen, Firat, Bapna, Johnson, Macherey, Foster,
  Jones, Schuster, Shazeer, Parmar, Vaswani, Uszkoreit, Kaiser, Chen, Wu, and
  Hughes}]{Chen:2018:ACL}
Mia~Xu Chen, Orhan Firat, Ankur Bapna, Melvin Johnson, Wolfgang Macherey,
  George Foster, Llion Jones, Mike Schuster, Noam Shazeer, Niki Parmar, Ashish
  Vaswani, Jakob Uszkoreit, Lukasz Kaiser, Zhifeng Chen, Yonghui Wu, and
  Macduff Hughes. 2018.
\newblock The best of both worlds: Combining recent advances in neural machine
  translation.
\newblock In \emph{ACL}.

\bibitem[{Cho et~al.(2014)Cho, Van~Merri{\"e}nboer, Gulcehre, Bahdanau,
  Bougares, Schwenk, and Bengio}]{cho2014learning}
Kyunghyun Cho, Bart Van~Merri{\"e}nboer, Caglar Gulcehre, Dzmitry Bahdanau,
  Fethi Bougares, Holger Schwenk, and Yoshua Bengio. 2014.
\newblock Learning phrase representations using {RNN} encoder-decoder for
  statistical machine translation.
\newblock In \emph{EMNLP}.

\bibitem[{Chung et~al.(2017)Chung, Ahn, and Bengio}]{Chung:2017:ICLR}
JunYoung Chung, Sungjin Ahn, and Yoshua Bengio. 2017.
\newblock {Hierarchical Multiscale Recurrent Neural Networks}.
\newblock In \emph{ICLR}.

\bibitem[{Conneau et~al.(2018)Conneau, Kruszewski, Lample, Barrault, and
  Baroni}]{conneau2018you}
Alexis Conneau, German Kruszewski, Guillaume Lample, Lo{\"\i}c Barrault, and
  Marco Baroni. 2018.
\newblock What you can cram into a single ${\$}{\&}!{\#}*$ vector: Probing
  sentence embeddings for linguistic properties.
\newblock In \emph{ACL}.

\bibitem[{Dehghani et~al.(2019)Dehghani, Gouws, Vinyals, Uszkoreit, and
  Kaiser}]{Dehghani:2019:ICLR}
Mostafa Dehghani, Stephan Gouws, Oriol Vinyals, Jakob Uszkoreit, and Lukasz
  Kaiser. 2019.
\newblock Universal transformers.
\newblock In \emph{ICLR}.

\bibitem[{Dou et~al.(2018)Dou, Tu, Wang, Shi, and Zhang}]{Dou:2018:EMNLP}
Ziyi Dou, Zhaopeng Tu, Xing Wang, Shuming Shi, and Tong Zhang. 2018.
\newblock Exploiting deep representations for neural machine translation.
\newblock In \emph{EMNLP}.

\bibitem[{Dou et~al.(2019)Dou, Tu, Wang, Wang, Shi, and Zhang}]{Dou:2019:AAAI}
Ziyi Dou, Zhaopeng Tu, Xing Wang, Longyue Wang, Shuming Shi, and Tong Zhang.
  2019.
\newblock Dynamic layer aggregation for neural machine translation.
\newblock In \emph{AAAI}.

\bibitem[{Hassan et~al.(2018)Hassan, Aue, Chen, Chowdhary, Clark, Federmann,
  Huang, Junczys-Dowmunt, Lewis, Li et~al.}]{hassan2018achieving}
Hany Hassan, Anthony Aue, Chang Chen, Vishal Chowdhary, Jonathan Clark,
  Christian Federmann, Xuedong Huang, Marcin Junczys-Dowmunt, William Lewis,
  Mu~Li, et~al. 2018.
\newblock Achieving human parity on automatic chinese to english news
  translation.
\newblock \emph{arXiv preprint arXiv:1803.05567}.

\bibitem[{He et~al.(2016)He, Zhang, Ren, and Sun}]{he2016deep}
Kaiming He, Xiangyu Zhang, Shaoqing Ren, and Jian Sun. 2016.
\newblock Deep residual learning for image recognition.
\newblock In \emph{CVPR}.

\bibitem[{Hochreiter and Schmidhuber(1997)}]{hochreiter1997long}
Sepp Hochreiter and J{\"u}rgen Schmidhuber. 1997.
\newblock Long short-term memory.
\newblock \emph{Neural computation}, 9(8):1735--1780.

\bibitem[{Koehn et~al.(2003)Koehn, Och, and Marcu}]{koehn2003statistical}
Philipp Koehn, Franz~Josef Och, and Daniel Marcu. 2003.
\newblock Statistical phrase-based translation.
\newblock In \emph{ACL}.

\bibitem[{Li et~al.(2018)Li, Tu, Yang, Lyu, and Zhang}]{Li:2018:EMNLP}
Jian Li, Zhaopeng Tu, Baosong Yang, Michael~R. Lyu, and Tong Zhang. 2018.
\newblock {Multi-Head Attention with Disagreement Regularization}.
\newblock In \emph{EMNLP}.

\bibitem[{Papineni et~al.(2002)Papineni, Roukos, Ward, and
  Zhu}]{papineni2002bleu}
Kishore Papineni, Salim Roukos, Todd Ward, and Wei-Jing Zhu. 2002.
\newblock Bleu: a method for automatic evaluation of machine translation.
\newblock In \emph{ACL}.

\bibitem[{Peters et~al.(2018)Peters, Neumann, Iyyer, Gardner, Clark, Lee, and
  Zettlemoyer}]{Peters:2018:NAACL}
Matthew~E. Peters, Mark Neumann, Mohit Iyyer, Matt Gardner, Christopher Clark,
  Kenton Lee, and Luke Zettlemoyer. 2018.
\newblock {Deep contextualized word representations}.
\newblock In \emph{NAACL}.

\bibitem[{Schuster and Paliwal(1997)}]{Schuster:1997:TSP}
Mike Schuster and Kuldip~K Paliwal. 1997.
\newblock Bidirectional recurrent neural networks.
\newblock \emph{IEEE Transactions on Signal Processing}, 45(11):2673--2681.

\bibitem[{Sennrich et~al.(2016)Sennrich, Haddow, and
  Birch}]{sennrich2016neural}
Rico Sennrich, Barry Haddow, and Alexandra Birch. 2016.
\newblock Neural machine translation of rare words with subword units.
\newblock In \emph{ACL}.

\bibitem[{Shaw et~al.(2018)Shaw, Uszkoreit, and Vaswani}]{Shaw:2018:NAACL}
Peter Shaw, Jakob Uszkoreit, and Ashish Vaswani. 2018.
\newblock {Self-Attention with Relative Position Representations}.
\newblock In \emph{NAACL}.

\bibitem[{Shen et~al.(2018)Shen, Zhou, Long, Jiang, Pan, and
  Zhang}]{Shen:2018:AAAI}
Tao Shen, Tianyi Zhou, Guodong Long, Jing Jiang, Shirui Pan, and Chengqi Zhang.
  2018.
\newblock {DiSAN: directional self-attention network for RNN/CNN-free language
  understanding}.
\newblock In \emph{AAAI}.

\bibitem[{Shen et~al.(2019)Shen, Tan, Sordoni, and Courville}]{Shen:2019:ICLR}
Yikang Shen, Shawn Tan, Alessandro Sordoni, and Aaron Courville. 2019.
\newblock Ordered neurons: Integrating tree structures into recurrent neural
  networks.
\newblock In \emph{ICLR}.

\bibitem[{Strubell et~al.(2018)Strubell, Verga, Andor, Weiss, and
  McCallum}]{Strubell:2018:EMNLP}
Emma Strubell, Patrick Verga, Daniel Andor, David Weiss, and Andrew McCallum.
  2018.
\newblock {Linguistically-Informed Self-Attention for Semantic Role Labeling}.
\newblock In \emph{EMNLP}.

\bibitem[{Sutskever et~al.(2014)Sutskever, Vinyals, and
  Le}]{sutskever2014sequence}
Ilya Sutskever, Oriol Vinyals, and Quoc~V Le. 2014.
\newblock Sequence to sequence learning with neural networks.
\newblock In \emph{NIPS}.

\bibitem[{Tran et~al.(2016)Tran, Bisazza, and Monz}]{Tran:2016:NAACL}
Ke~Tran, Arianna Bisazza, and Christof Monz. 2016.
\newblock {Recurrent memory networks for language modeling}.
\newblock In \emph{NAACL}.

\bibitem[{Tran et~al.(2018)Tran, Bisazza, and Monz}]{Tran:2018:arXiv}
Ke~Tran, Arianna Bisazza, and Christof Monz. 2018.
\newblock The importance of being recurrent for modeling hierarchical
  structure.
\newblock In \emph{EMNLP}.

\bibitem[{Tu et~al.(2017)Tu, Liu, Lu, Liu, and Li}]{Tu:2017:TACL}
Zhaopeng Tu, Yang Liu, Zhengdong Lu, Xiaohua Liu, and Hang Li. 2017.
\newblock {Context gates for neural machine translation}.
\newblock \emph{TACL}.

\bibitem[{Vaswani et~al.(2017)Vaswani, Shazeer, Parmar, Uszkoreit, Jones,
  Gomez, Kaiser, and Polosukhin}]{Vaswani:2017:NIPS}
Ashish Vaswani, Noam Shazeer, Niki Parmar, Jakob Uszkoreit, Llion Jones,
  Aidan~N Gomez, {\L}ukasz Kaiser, and Illia Polosukhin. 2017.
\newblock Attention is all you need.
\newblock In \emph{NIPS}.

\bibitem[{Voita et~al.(2018)Voita, Serdyukov, Sennrich, and
  Titov}]{Voita:2018:ACL}
Elena Voita, Pavel Serdyukov, Rico Sennrich, and Ivan Titov. 2018.
\newblock Context-aware neural machine translation learns anaphora resolution.
\newblock In \emph{ACL}.

\bibitem[{Wang et~al.(2018)Wang, Li, Xiao, Li, Li, and Zhu}]{wang:2018:COLING}
Qiang Wang, Fuxue Li, Tong Xiao, Yanyang Li, Yinqiao Li, and Jingbo Zhu. 2018.
\newblock Multi-layer representation fusion for neural machine translation.
\newblock In \emph{COLING}.

\bibitem[{Yang et~al.(2019)Yang, Li, Wong, Chao, Wang, and Tu}]{Yang:2019:AAAI}
Baosong Yang, Jian Li, Derek~F. Wong, Lidia~S. Chao, Xing Wang, and Zhaopeng
  Tu. 2019.
\newblock Context-aware self-attention networks.
\newblock In \emph{AAAI}.

\bibitem[{Yang et~al.(2016)Yang, Yuan, Wu, Salakhutdinov, and
  Cohen}]{Yang:2016:NIPS}
Zhilin Yang, Ye~Yuan, Yuexin Wu, Ruslan Salakhutdinov, and William~W Cohen.
  2016.
\newblock {Review networks for caption generation}.
\newblock In \emph{NIPS}.

\end{thebibliography}
\bibliographystyle{acl_natbib}

\end{document}